\documentclass[conference,letterpaper]{ieeeconf}
\IEEEoverridecommandlockouts    

\usepackage[utf8]{inputenc}
\usepackage[T1]{fontenc}
\usepackage{amsmath,amssymb,amsfonts}
\usepackage{graphicx}
\usepackage{balance}
\usepackage{booktabs}
\usepackage{multirow}
\usepackage{xcolor}
\usepackage{url}
\usepackage{cite}
\usepackage{array}
\usepackage{tabularx}
\let\labelindent\relax
\usepackage{enumitem}
\usepackage{algorithm}
\usepackage{algpseudocode}
\usepackage{subcaption}
\usepackage{caption}
\usepackage{tikz}
\usetikzlibrary{arrows.meta, positioning, shapes.geometric,
                fit, backgrounds, calc, decorations.pathreplacing}

\definecolor{threadS}{RGB}{31,119,180}   
\definecolor{threadQ}{RGB}{255,127,14}   
\definecolor{threadE}{RGB}{44,160,44}    
\definecolor{highlight}{RGB}{214,39,40}  

\usepackage{scalerel}
\usepackage{tikz}
\usetikzlibrary{svg.path}

\definecolor{orcidlogocol}{HTML}{A6CE39}
\tikzset{
	orcidlogo/.pic={
			\fill[orcidlogocol] svg{M256,128c0,70.7-57.3,128-128,128C57.3,256,0,198.7,0,128C0,57.3,57.3,0,128,0C198.7,0,256,57.3,256,128z};
			\fill[white] svg{M86.3,186.2H70.9V79.1h15.4v48.4V186.2z}
			svg{M108.9,79.1h41.6c39.6,0,57,28.3,57,53.6c0,27.5-21.5,53.6-56.8,53.6h-41.8V79.1z M124.3,172.4h24.5c34.9,0,42.9-26.5,42.9-39.7c0-21.5-13.7-39.7-43.7-39.7h-23.7V172.4z}
			svg{M88.7,56.8c0,5.5-4.5,10.1-10.1,10.1c-5.6,0-10.1-4.6-10.1-10.1c0-5.6,4.5-10.1,10.1-10.1C84.2,46.7,88.7,51.3,88.7,56.8z};
		}
}

\newcommand\orcidicon[1]{\href{https://orcid.org/#1}{\mbox{\scalerel*{
				\begin{tikzpicture}[yscale=-1,transform shape]
					\pic{orcidlogo};
				\end{tikzpicture}
			}{|}}}}

\usepackage{hyperref}
\usepackage{booktabs}

\newcommand{\etal}{\textit{et~al.}}

\begin{document}

\title{\LARGE \bf CADENet: Condition-Adaptive Asynchronous Dual-Stream Enhancement Network for Adverse Weather Perception in Autonomous Driving}

\author{Sherif~Khairy$^{1,2}$, and Catherine~M.~Elias$^{1,2\orcidicon{0000-0002-1444-9816}\,}$,~\IEEEmembership{Member,~IEEE} 
	\thanks{*This work was not supported by any organization}%
	\thanks{$^{1}$Computer Science \& Engineering Department, German University in Cairo (GUC), Egypt {\tt\small Sherif.ammen@gmail.com,  catherine.elias@ieee.org} }%
	\thanks{$^{2}$C-DRiVeS Lab: Cognitive Driving Research in Vehicular Systems, Cairo, Egypt {\tt\small cdrives.researchlab@gmail.com}}%
}

\markboth{Journal of \LaTeX\ Class Files,~Vol.~14, No.~8, August~2015}%
{author1 \MakeLowercase{\textit{et al.}}:title here}
%



\maketitle

\begin{abstract}
Adverse weather---rain, fog, sand, and snow---degrades camera-based object detection in autonomous vehicles. Existing enhancement-then-detect approaches stall the safety-critical perception loop, violating hard real-time requirements. Progress on this problem is also constrained by an under-recognised \emph{evaluation ceiling}: ground truth annotated on degraded images cannot credit a detector that recovers objects the annotators themselves could not see, so a genuinely useful enhancement can register as a near-flat $F_1$ gain. The paper presents \textbf{CADENet} (Condition-Adaptive asynchronous Dual-stream Enhancement Network), a training-free three-thread system: Thread~S (YOLOv11n) delivers detections at full frame rate with \textbf{zero added latency}; Thread~Q applies condition-adaptive enhancement (CAPE) and fuses results via entropy-guided NMS (EG-NMS) without blocking Thread~S; Thread~E provides \textbf{CLIP zero-shot weather classification}, so \textbf{new weather categories require only a new text prompt}---no labelled data, no retraining. Evaluated on 1,327 DAWN images (YOLOv11m, IoU@0.5, conf\,=\,0.25), CADENet achieves $\Delta\text{Recall}\!=\!{+}0.0103$ (micro), $\Delta F_1\!=\!{+}0.0230$ on snow and $\Delta F_1\!=\!{+}0.0038$ on rain. We formalise the annotation completeness bias on DAWN-class data, so the reported $F_1$ values are \emph{lower bounds} on the true gain; recall is the annotation-gap-immune headline metric. Thread~S sustains ${\sim}44$\,FPS regardless of enhancement load. No model retraining or additional sensor hardware is required.
\end{abstract}


\section{Introduction}

Camera-based object detection is the dominant perception modality in
autonomous vehicles, but degrades under adverse weather: rain streaks
corrupt feature maps, fog attenuates edges, sand scatters light, and snow introduces specular patches---each reducing recall and elevating
false-positive rates.

Prior work addresses weather robustness along three main axes.
\textit{Data augmentation}~\cite{michaelis2019benchmarking} retrains
detectors on synthetic corruptions, but requires large retraining cost
and degrades on unseen severity levels.
\textit{Restoration-then-detection}~\cite{liu2022iayolo,qin2022denet,chen2025awdyolo} cascades a restoration network with a standard detector,
achieving strong benchmark numbers at the expense of paired training
data and a single blocking pipeline.
\textit{Multi-modal sensor fusion}~\cite{bijelic2020seeingfog} fuses
LiDAR, radar, and cameras but demands calibrated multi-sensor rigs.

\textbf{The deployment gap.}
All three axes block the safety-critical output path for 90--160\,ms,
violating hard real-time requirements---a constraint unmodelled by
IA-YOLO~\cite{liu2022iayolo}, DENet~\cite{qin2022denet}, and
AWD-YOLO~\cite{chen2025awdyolo}.

\textbf{Our approach.}
\textbf{CADENet} separates the safety path (Thread~S, YOLOv11n,
${\sim}44$\,FPS, \textbf{zero added latency}) from a parallel
quality-enhancement path (Thread~Q: WEM~$\to$~CAPE~$\to$~YOLOv11m~$\to$~EG-NMS)
and a background analytics thread (Thread~E: CLIP zero-shot weather
classification~\cite{radford2021clip}\,+\,ResNet50 $k$-NN filter
recommendation).

\textbf{Evaluation note.}
DAWN GT~\cite{kenk2020dawn} was annotated on degraded images;
post-enhancement detections of newly visible objects are penalised as FP,
so \textbf{reported $\Delta F_1$ is a lower bound on true gain};
recall is annotation-gap-immune (formal treatment:
Section~\ref{sec:annotation_bias}).

\subsection*{Contributions}
\begin{enumerate}
  \item \textbf{Three-thread asynchronous architecture}:
  Thread~S (YOLOv11n, ${\sim}23$\,ms, ${\sim}44$\,FPS GPU) and Thread~Q
  (CAPE\,+\,YOLOv11m, ${\sim}90$--$160$\,ms) run concurrently;
  Thread~Q injects $k$-step Kalman-projected detections without blocking
  Thread~S.
  \item \textbf{Zero-shot extensible weather detection (CLIP)}:
  Thread~E runs CLIP (ViT-B/32)~\cite{radford2021clip}; Thread~Q's
  heuristic WEM reads CLIP's label when its confidence
  spread $\Delta_\text{top2}<0.15$.
  \textbf{New weather categories require only a new text prompt}---no
  retraining, no labelled data.
  \item \textbf{Modular CAPE}: training-free, severity-continuous
  filters for all four conditions (rain: 5-stage morphological derain;
  fog: DCP~\cite{he2011dcp}; sand/snow: CLAHE); each branch
  independently replaceable with no detector retraining.
  \item \textbf{PEE\,+\,EG-NMS}: vectorised Shannon entropy over
  $16\!\times\!16$ patches yields reliability $R(j)\!\in\![0,1]$;
  EG-NMS weights both stream detections by $R(j)$,
  \textbf{suppressing weather-induced YOLO false positives} with no
  learnt parameters.
  \item \textbf{Annotation-bias formalisation}: $\Delta F_1$ on DAWN
  is a lower bound on true gain; recall is the annotation-gap-immune
  primary metric, applicable to any enhancement method on DAWN-class
  datasets.
\end{enumerate}
\section{Related Work}

\subsection{Physics-Based Image Restoration}

He~\etal~\cite{he2011dcp} established the \emph{Dark Channel Prior}
(DCP): at least one colour channel is near-zero in any clear patch,
while haze raises the minimum intensity.
DCP fails under artificial light sources where lamp returns contaminate
atmospheric light $A$ (observed: fog $\Delta F_1=-0.0078$,
Section~\ref{sec:results}).
Deep networks such as AOD-Net~\cite{li2017aodnet} achieve higher PSNR
but require paired training data.
CADENet retains DCP for its $<\!30$\,ms runtime.

Yang~\etal~\cite{yang2017rain} introduced deep joint rain detection-removal
at the expense of synthetic training data;
CADENet's CAPE uses a training-free 5-stage morphological derain.

Snow and sand share a common structure: reflective or scattering
particulates reduce local contrast.
DesnowNet~\cite{liu2018desnow} and TransWeather~\cite{valanarasu2022transweather}
address snow and multi-weather restoration with learned networks.
CADENet applies CLAHE on the L channel (clip\,=\,2.0, tile\,=\,$8\!\times\!8$),
achieving $\Delta F_1 = +0.0230$ on snow with no training cost.

\subsection{Joint Enhancement and Detection}

IA-YOLO~\cite{liu2022iayolo} introduced a differentiable image
processing (DIP) module whose parameters are predicted by a small CNN,
jointly optimised end-to-end with YOLO.
DENet~\cite{qin2022denet} cascades a U-Net restoration network with
detection under joint supervision.
AWD-YOLO~\cite{chen2025awdyolo}---the strongest published DAWN
baseline---adds a Faster Local-Region Self-Attention (FLRSA) module to
YOLOv11, reporting mAP@0.5\,=\,67.7\% on DAWN.

\textbf{Shared limitation}: all present \emph{single-thread blocking
pipelines}---the safety-critical output path stalls during enhancement.
CADENet is the first to formalise the safety/quality separation as an
explicit architectural primitive for single-camera AV systems.

\subsection{Multi-Modal Sensor Fusion and Entropy Gating}

Bijelic~\etal~\cite{bijelic2020seeingfog} compute per-sensor
signal-quality entropy to gate fusion between gated NIR, LiDAR, radar,
and colour camera under dense fog.
This directly motivates PEE and EG-NMS: $R(j)$ is the single-camera
analogue---pixel-level reliability within one image, no extra hardware.

\subsection{Tracking and Temporal Coherence}

SORT~\cite{bewley2016sort} establishes the Kalman+Hungarian paradigm
that CADENet's KTT module extends.
UncertaintyTrack~\cite{lee2024uncertaintytrack} motivates CADENet's
per-track confidence smoothing.
KTT extends SORT with $k$-step forward-projection for Thread~Q
detections arriving $k$ frames late.

\section{CADENet Architecture}
\label{sec:architecture}

\subsection{Problem Formulation}

Let $F_t \in \mathbb{R}^{H\times W\times 3}$ be an RGB frame captured
at camera period $T_\text{cam}$ (e.g., 33\,ms at 30\,Hz).
Weather condition $w\in\mathcal{W}=\{\text{rain, fog, sand, snow,
clear}\}$ and severity $s\in[0,1]$ are latent variables estimated
from $F_t$.
The system maintains a live track state
$\mathcal{T}_t = \{(\mathbf{b}_i, c_i, \bar{\sigma}_i, \tau_i,
\hat{\mathbf{x}}_i)\}_{i=1}^{N}$,
where $\mathbf{b}_i$ is a bounding box, $c_i$ is class,
$\bar{\sigma}_i \in [0,1]$ is smoothed confidence, $\tau_i$ is a
track ID, and $\hat{\mathbf{x}}_i$ is the Kalman state vector.

The system must satisfy two simultaneous constraints:
\begin{itemize}
  \item \textbf{Hard real-time}: $\mathcal{T}_t$ is updated every
        $T_\text{cam}$ with latency $\Delta t_S \le T_\text{cam}$
        (Thread~S constraint).
  \item \textbf{Quality refinement}: Thread~Q produces enhanced
        detections $D_{t-k}$ at time $t$, absorbed into
        $\mathcal{T}_t$ via $k$-step Kalman projection, without
        blocking Thread~S.
\end{itemize}

\subsection{Three-Thread Architecture}

\begin{figure*}[t]
\centering
\begin{tikzpicture}[
  font=\small,
  tbox/.style={draw, rounded corners=3pt, minimum width=3.6cm,
               minimum height=0.65cm, align=center, inner sep=5pt,
               text width=3.2cm},
  sbox/.style={tbox, fill=threadS!10, draw=threadS!70, thick},
  qbox/.style={tbox, fill=threadQ!10, draw=threadQ!70, thick},
  ebox/.style={tbox, fill=threadE!10, draw=threadE!70, thick},
  hbox/.style={draw, rounded corners=4pt, minimum width=4.0cm,
               minimum height=0.75cm, align=center,
               font=\small\bfseries, inner sep=5pt},
  shbox/.style={hbox, fill=threadS!30, draw=threadS},
  qhbox/.style={hbox, fill=threadQ!30, draw=threadQ},
  ehbox/.style={hbox, fill=threadE!30, draw=threadE},
  arr/.style={-Stealth, thick},
  sarr/.style={arr, color=threadS!80!black},
  qarr/.style={arr, color=threadQ!80!black},
  earr/.style={arr, color=threadE!80!black},
]
\def\xs{-6.2cm}
\def\xq{0cm}
\def\xe{6.2cm}

\node[draw=gray!60, fill=gray!10, rounded corners=4pt,
      minimum width=2.4cm, minimum height=0.65cm, align=center]
      (cam) {\textbf{Camera} $F_t$};

\node[shbox, below=0.9cm of cam, xshift=\xs] (sh)
     {THREAD~S\\Safety-Critical};
\node[sbox, below=0.35cm of sh] (s1)
     {YOLOv11n\quad$\sim\!23$\,ms GPU};
\node[sbox, below=0.3cm of s1]  (ktt)
     {KTT: Kalman predict+assign};
\node[sbox, below=0.3cm of ktt] (out)
     {$\mathcal{T}_t$\ AV Output\quad$\sim\!44$\,FPS};

\node[qhbox, below=0.9cm of cam, xshift=\xq] (qh)
     {THREAD~Q\\Quality-Enhancement};
\node[qbox, below=0.35cm of qh]  (wem)
     {M1: WEM\\\textit{weather + severity}};
\node[qbox, below=0.3cm of wem]  (pee)
     {M2: PEE\\\textit{reliability map} $R(j)$};
\node[qbox, below=0.3cm of pee]  (cape)
     {M3: CAPE\quad$\sim\!30$--$50$\,ms\\\textit{enhancement filter}};
\node[qbox, below=0.3cm of cape] (ym)
     {M4: YOLOv11m\quad$\sim\!35$\,ms GPU};
\node[qbox, below=0.3cm of ym]   (egnms)
     {M5: EG-NMS + Kalman proj.};

\node[ehbox, below=0.9cm of cam, xshift=\xe] (eh)
     {THREAD~E\\Experience Accum.};
\node[ebox, below=0.35cm of eh]   (clip)
     {M6a: CLIP (ViT-B/32)\\\textit{zero-shot weather}};
\node[ebox, below=0.3cm of clip]  (rn50)
     {M6b: ResNet50\quad$\sim\!68$\,ms CPU};
\node[ebox, below=0.3cm of rn50]  (knn)
     {M6c: $k$-NN / SED};
\node[ebox, below=0.3cm of knn]   (sed)
     {M6d: DB update (online)};

\draw[arr,gray!70] (cam.south) -- ++(0,-0.45cm) -| (sh.north);
\draw[arr,gray!70] (cam.south) -- (qh.north);
\draw[arr,gray!70] (cam.south) -- ++(0,-0.45cm) -| (eh.north);

\draw[sarr](sh)--(s1); \draw[sarr](s1)--(ktt); \draw[sarr](ktt)--(out);

\draw[qarr](qh)--(wem); \draw[qarr](wem)--(pee); \draw[qarr](pee)--(cape);
\draw[qarr](cape)--(ym); \draw[qarr](ym)--(egnms);

\draw[earr](eh)--(clip); \draw[earr](clip)--(rn50);
\draw[earr](rn50)--(knn); \draw[earr](knn)--(sed);

\draw[earr, dashed, thick] (knn.west) --
     node[above, sloped, font=\footnotesize]{\texttt{rec*}} (cape.east);

\draw[qarr, dashed, thick] (egnms.west) -|
     node[pos=0.12, above, font=\footnotesize]{inject $k$ steps}
     (ktt.south);

\draw[earr, dashed, thick] (egnms.east) -|
     node[pos=0.12, right, font=\footnotesize]{proxy $\hat{q}$}
     (sed.south);

\begin{scope}[on background layer]
  \node[fill=threadS!5, rounded corners=8pt, draw=threadS!25, dashed,
        inner sep=6pt, fit=(sh)(s1)(ktt)(out)] {};
  \node[fill=threadQ!5, rounded corners=8pt, draw=threadQ!25, dashed,
        inner sep=6pt, fit=(qh)(wem)(pee)(cape)(ym)(egnms)] {};
  \node[fill=threadE!5, rounded corners=8pt, draw=threadE!25, dashed,
        inner sep=6pt, fit=(eh)(clip)(rn50)(knn)(sed)] {};
\end{scope}

\end{tikzpicture}
\caption{CADENet three-thread architecture.
\textbf{S} (blue): YOLOv11n at $\sim\!44$\,FPS, zero added latency.
\textbf{Q} (orange): WEM\,$\to$\,PEE\,$\to$\,CAPE\,$\to$\,YOLOv11m\,$\to$\,EG-NMS;
injects $k$-step Kalman-projected detections into the live track state.
\textbf{E} (green): CLIP zero-shot classification and ResNet50 $k$-NN
filter recommendation written to a lock-free atomic slot at zero cost.
Dashed arrows: asynchronous cross-thread communication.}
\label{fig:arch}
\end{figure*}
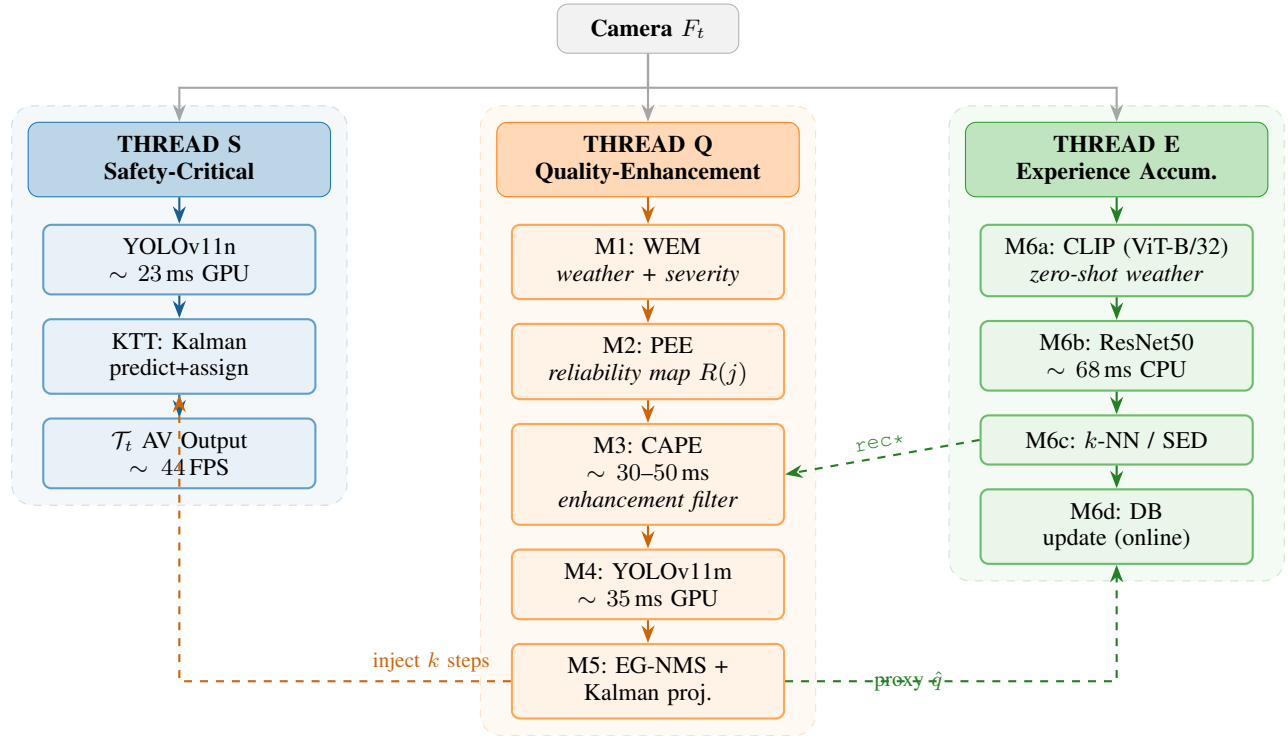

Fig.~\ref{fig:arch} illustrates the top-level architecture.
Three OS-level threads operate concurrently on each incoming frame.

\textbf{Thread~S} (synchronous, safety-critical) runs YOLOv11n~\cite{wang2024yolov11} on
frame $F_t$ and updates $\mathcal{T}_t$ with the Hungarian-assigned
Kalman update.
Latency is ${\sim}23$\,ms GPU (${\sim}44$\,FPS on RTX\,3050 Laptop);
Thread~S never waits for any other thread.

\textbf{Thread~Q} (asynchronous, quality-enhancement) processes the
same frame through WEM $\rightarrow$ PEE $\rightarrow$ CAPE
$\rightarrow$ YOLOv11m $\rightarrow$ EG-NMS.
The resulting fused detections are Kalman-projected $k$ steps
forward and injected into $\mathcal{T}_t$ via mutex write.
Thread~Q completes in ${\sim}90$--$160$\,ms (CPU+GPU, condition-dependent:
CLAHE conditions ${\sim}90$\,ms, DCP/derain ${\sim}160$\,ms;
$k\!=\!\lceil\Delta t_Q/T_\text{cam}\rceil \approx 3$--5 frames at 30\,Hz).

\textbf{Thread~E} (background, lowest priority) decouples
CLIP weather classification (${\sim}39$\,ms GPU), ResNet50 scene
embedding (${\sim}68$\,ms CPU), and $k$-NN filter lookup from Thread~Q's latency budget.
Thread~E writes to a \emph{lock-free atomic slot}; Thread~Q reads
it at CAPE entry with zero synchronisation cost (WEM's own estimate
is used if the slot is empty).
Thread~E is \textbf{horizontally scalable}: new analytics modules are
added without modifying Thread~S or Thread~Q.

\subsection{Module 1: Weather Estimation Module (WEM)}
\label{sec:wem}

WEM classifies condition $w \in \mathcal{W}$ and severity $s$
from a single frame via lightweight heuristic features, then
selects the corresponding CAPE filter branch.
When heuristic confidence is ambiguous, WEM reads Thread~E's
CLIP-quality label from the lock-free atomic slot, delivering
high-accuracy weather classification with zero impact on
Thread~Q's latency budget.

\textbf{Heuristic classification (real-time path).}
Five features on LAB/HSV colour spaces---L-channel mean $\mu_L$,
std $\sigma_L$, S-channel mean $\mu_S$, Canny edge density
$\rho_e$, vertical-edge ratio $r_v$---drive lightweight threshold
rules:
\begin{align*}
  \text{fog}  &: \sigma_L < 35 \;\wedge\; \rho_e < 0.1  \\
  \text{rain} &: r_v > 3.0   \;\wedge\; \mu_S < 60  \\
  \text{haze} &: \mu_S < 40  \;\wedge\; \sigma_L < 45
\end{align*}
Severity $s$ is derived from the magnitude of the leading feature
(e.g., $1 - \sigma_L/35$ for fog, clamped to $[0,1]$).
Top-2 probability spread $\Delta_\text{top2}$ quantifies
classification confidence.

\textbf{CLIP disambiguation and filter routing.}
If $\Delta_\text{top2} < 0.15$, WEM reads CLIP label $w_\text{CLIP}$
from Thread~E's atomic slot at zero latency cost; the resolved $w$
selects the CAPE branch (DCP / derain / CLAHE) and biases parameters
via Thread~E's $k$-NN recommendation $\text{rec}^*$.

\subsection{Module 2: Patch-Level Entropy Estimation (PEE)}

PEE quantifies per-region quality via Shannon entropy over
$16{\times}16$ patches:
\begin{equation}
  H(j) = -\sum_{k=0}^{15} h_j(k)\log_2 h_j(k), \quad
  R(j) = 1 - \frac{H(j)}{\log_2 16}
\end{equation}
where $h_j$ is the 16-bin normalised histogram of patch $j$ and
$R(j)\in[0,1]$ is the reliability score.
High $R(j)$ indicates a sharp, information-rich patch; low $R(j)$
indicates a degraded or uniform patch.
The reliability map is bilinearly upsampled to full resolution for
use in EG-NMS.

\subsection{Module 3: Condition-Adaptive Parameterised Enhancement (CAPE)}
\label{sec:cape}

CAPE selects a training-free enhancement filter per estimated condition,
with parameters that vary continuously with severity $s$;
each branch is independently replaceable.

\subsubsection{Rain: 5-Stage Morphological Derain}

\textit{Stage~1 --- streak detection}: a median-subtracted
difference image is thresholded and morphologically opened with a
$1\!\times\!7$ vertical kernel to isolate streak-shaped regions.
The rain pixel fraction $\rho_\text{rain}$ governs subsequent stages.

\textit{Stage~2 --- selective inpainting/blend}: if
$0.001 < \rho_\text{rain} < 0.30$, the streak mask is inpainted
with TELEA~\cite{telea2004inpaint}; for $\rho_\text{rain}\!\ge\!0.30$
(heavy rain), a median-blend fallback is used to avoid mask-scale
artefacts.

\textit{Stage~3 --- conditional gamma}: brightness correction
$\gamma = \text{clamp}(130/\max(\mu_L,30),\,1.05,\,1.40)$ is
applied only if $\mu_L < 130$, preventing over-brightening of
already well-exposed frames.

\textit{Stages~4--5}: L-channel CLAHE (clip\,=\,1.5) and bilateral
smoothing ($d\!=\!5$) suppress residual noise.

\subsubsection{Fog: Dark Channel Prior}

The DCP atmospheric scattering model $I^c(x) = J^c(x)T(x) +
A^c(1{-}T(x))$ is inverted per~\cite{he2011dcp}:
\begin{equation}
  T(x) = 1 - \alpha\cdot\!\min_{y\in\Omega(x)}\min_c
         \frac{I^c(y)}{A^c}, \quad
  \alpha = 0.5 + 0.4s
\end{equation}
where $\Omega(x)$ is a $15\!\times\!15$ local patch and
$A$ is the mean intensity of the top-0.1\% pixels ranked by dark
channel value.
$T(x)$ is clamped to $[0.1,\,1.0]$ to prevent division instability.
Post-DCP CLAHE (clip\,=\,2.0) restores micro-contrast.

\textit{Failure mode}: nighttime lamp returns contaminate $A$,
over-darkening the scene ($\Delta F_1=-0.0078$ on fog;
brightness gate fix: Section~\ref{sec:conclusion}).

\subsubsection{Sand and Snow: CLAHE}

Both conditions attenuate local contrast via scattering.
CLAHE with clip\,=\,2.0 applied to the LAB L channel compensates
without hue distortion, requiring no training data.

\subsection{Module 4: Entropy-Guided NMS (EG-NMS)}

Detections from Thread~S (YOLOv11n on $F_t$) and Thread~Q
(YOLOv11m on enhanced $\hat{F}_{t-k}$) are fused using patch
reliability:
\begin{align}
  d_S.\text{score} &= R(j_d)\cdot d_S.\text{conf}, \label{eq:egs}\\
  d_Q.\text{score} &= (1{-}R(j_d))\cdot d_Q.\text{conf}. \label{eq:egq}
\end{align}
High $R(j_d)$ (clear patch) up-weights Thread~S~\eqref{eq:egs};
low $R(j_d)$ (degraded patch) up-weights Thread~Q~\eqref{eq:egq},
where $j_d$ indexes the patch containing detection $d$.
Standard NMS ($\text{IoU}=0.45$) is applied to the pooled candidate set.
The fused result $D_{t-k}$ is Kalman-projected $k$ steps forward and
injected into $\mathcal{T}_t$.

\subsection{Module 5: Kalman Temporal Tracker (KTT)}

KTT adapts SORT~\cite{bewley2016sort} for the asynchronous dual-stream
context.
State vector: $\mathbf{x}_i=[c_x,c_y,a,h,\dot{c}_x,\dot{c}_y,\dot{a}]^\top$
with constant-velocity transition.
Hungarian assignment uses $C_{ij}=1-\text{IoU}(\text{track}_i,\text{det}_j)$
with threshold~0.3; tracks born after 1 hit, killed after 3 missed frames.

\textbf{Async injection}: Thread~Q applies $k$ Kalman prediction steps
to each quality detection before Hungarian matching against the current
track state, ensuring temporal coherence despite the $k$-frame lag.
Confidence is smoothed per-track as $\bar{\sigma}_i = 0.7\sigma_\text{new}
+ 0.3\bar{\sigma}_{i,t-1}$, suppressing single-frame enhancement spikes.

\subsection{Module 6: Scene Embedding Database (Thread~E)}

Thread~E runs a sequential background pipeline on each buffered frame
$\hat{F}_{t-k}$, writing results to a lock-free atomic slot consumed by
Thread~Q without mutex overhead (\textit{release-acquire} ordering).

\textbf{M6a -- CLIP zero-shot weather classification.}
CLIP (ViT-B/32)~\cite{radford2021clip} encodes the frame and ranks
natural-language weather prompts via cosine similarity (${\sim}39$\,ms
GPU, background only).
Thread~Q reads the top label $w_\text{CLIP}$ when heuristic confidence
is ambiguous ($\Delta_\text{top2}<0.15$) at zero latency cost.
New conditions require only a new text prompt---no fine-tuning.

\textbf{M6b/c -- ResNet50 scene embedding and $k$-NN filter recommendation.}
A ResNet50 (ImageNet) penultimate-layer embedding
$\mathbf{v}_{t-k}\in\mathbb{R}^{2048}$ (L2-normalised) is extracted.
The SED (Scene Embedding Database) is queried for $k\!=\!5$ nearest
neighbours by cosine similarity, filtered to entries with $\Delta F_1>0$,
and scored as:
\begin{equation}
  \text{score}(c) = \text{sim}(c)\cdot\exp(2\cdot c.\Delta F_1)
\end{equation}
The recommendation $\text{rec}^*=\arg\max_c \text{score}(c)$, together
with $w_\text{CLIP}$, is written to the atomic slot; Thread~Q reads it
at CAPE entry.
After each cycle, Thread~E appends the embedding, parameters, and
proxy quality signal (confidence delta) to the SED for online adaptation.
\section{Experimental Setup}
\label{sec:setup}

\subsection{Dataset and Protocol}

We evaluate on the DAWN dataset~\cite{kenk2020dawn}, the standard
adverse-weather benchmark for YOLO-class detection systems~\cite{chen2025awdyolo}.
We use all DAWN images with valid Pascal VOC annotations:
fog~600, rain~200, sand~323, snow~204 --- totalling 1,327 images
(1,325 with parseable GT). Fog comprises 45\% of the set, reflecting
DAWN's sampling distribution and dominating the micro-average.

All runs use YOLOv11m, confidence threshold\,=\,0.25, IoU@0.5.
We distinguish \textbf{C1} (YOLOv11m on the original degraded frame
vs.\ GT) from \textbf{C2} (YOLOv11m on the CAPE-enhanced frame
vs.\ the same GT), with $\Delta F_1 = F_1^\text{C2} - F_1^\text{C1}$
as the primary metric.
Per-image flags use a $\pm0.01$ deadband: Flag-0 ($|\Delta F_1|<0.01$),
Flag-1 (improved), Flag-2 (degraded).

\subsection{Implementation and Hardware}
\label{sec:impl}

All experiments were conducted on a laptop workstation with an
NVIDIA GeForce RTX~3050 6\,GB Laptop GPU (Compute~8.6, driver~560.94,
CUDA~12.6), a 12th-Gen Intel Core i5-12450H (8~cores), 8\,GB RAM,
running Windows~11.
The benchmark pipeline ran sequentially (single-threaded) for reproducibility.
All component latencies were measured directly on the test hardware
(GPU timings: 10 warmup\,+\,50 timed calls with \texttt{cuda.synchronize()};
CPU timings: 5 warmup\,+\,100 timed calls):
Thread~S YOLOv11n $23\!\pm\!12$\,ms GPU;
YOLOv11m $28\!\pm\!8$\,ms GPU;
CAPE fog DCP $80\!\pm\!21$\,ms CPU,
rain derain $64\!\pm\!29$\,ms CPU,
sand/snow CLAHE $10$--$20$\,ms CPU;
CLIP ViT-B/32 $39\!\pm\!13$\,ms GPU;
ResNet50 $68\!\pm\!6$\,ms CPU;
EG-NMS $0.3\!\pm\!0.1$\,ms.

\textbf{CAPE routing protocol.}
Images are routed via GT condition labels (\emph{upper-bound routing}),
isolating CAPE quality from WEM accuracy; reported $\Delta F_1$ reflects
maximum achievable gain---end-to-end performance additionally depends on
WEM or CLIP accuracy.

\textbf{WEM classification accuracy.}
WEM heuristic accuracy was evaluated independently on 1,027 labelled DAWN
images (Table~\ref{tab:wem}).
Overall accuracy is 40.4\%, with strong performance on snow (97.1\%)
but low accuracy on fog (8.7\%) and rain (20.5\%), caused by WEM's
LAB-space snow-bias under uniform-grey degraded scenes.
CLIP disambiguation (Module~6a) targets these failure cases, motivating
Thread~E's zero-cost high-accuracy classification.

\begin{table}[t]
\centering
\caption{WEM heuristic classification accuracy on DAWN GT labels,
evaluated independently of the C1/C2 benchmark (which uses GT routing).}
\label{tab:wem}
\renewcommand{\arraystretch}{1.05}
\setlength{\tabcolsep}{5pt}
\begin{tabular}{lrrrl}
\toprule
Cond.  & $N$ & Correct & Acc. & Primary error \\
\midrule
Snow   & 204 & 198 & 97.1\% & --- \\
Sand   & 323 & 150 & 46.4\% & predicted snow \\
Rain   & 200 &  41 & 20.5\% & predicted snow (77\%) \\
Fog    & 300 &  26 &  8.7\% & predicted snow (80\%) \\
\midrule
\textbf{Overall} & \textbf{1027} & \textbf{415} & \textbf{40.4\%} & \\
\bottomrule
\end{tabular}
\end{table}

\section{Results}
\label{sec:results}

\subsection{Per-Weather $F_1$ Breakdown}

Table~\ref{tab:weather} reports the main result.
Across all 1,325 GT images, CAPE achieves micro recall
$+0.0103$ (+101 TP, $-101$ FN) and macro $F_1$ $+0.0006$
($0.7166$ vs.\ $0.7160$ C1).
The micro $F_1$ is $-0.0005$ ($0.7015$ vs.\ $0.7020$), near-neutral
due to fog's 45\% weight.
All FP increase ($+199$) mixes enhancement artefacts with
annotation-gap detections --- separated only by manual re-annotation
(Section~\ref{sec:annotation_bias}).

\begin{table}[t]
\centering
\caption{Per-weather macro $F_1$ and flag distribution
(YOLOv11m, IoU@0.5, conf\,=\,0.25).
F0\,=\,unchanged ($|\Delta F_1|<0.01$);
F1\,=\,improved; F2\,=\,degraded.}
\label{tab:weather}
\renewcommand{\arraystretch}{1.1}
\setlength{\tabcolsep}{5.5pt}
\begin{tabular}{lrrrrrrr}
\toprule
Cond. & $N$ & C1 $F_1$ & C2 $F_1$ & $\Delta F_1$ & F0 & F1 & F2 \\
\midrule
Snow & 204 & 0.750 & 0.773 & $\mathbf{+0.023}$ & 103 & 59 & 42 \\
Rain & 200 & 0.732 & 0.736 & $\mathbf{+0.004}$ & 102 & 47 & 51 \\
Sand & 323 & 0.716 & 0.716 & \phantom{$+$}0.000 & 170 & 72 & 81 \\
Fog  & 600 & 0.699 & 0.692 & $-0.008$           & 270 & 156 & 174 \\
\midrule
\textbf{Macro} & \textbf{1327} & \textbf{0.716} & \textbf{0.717}
               & $\mathbf{+0.001}$ & 645 & 334 & 348 \\
\bottomrule
\end{tabular}
\end{table}

\textbf{Snow} (+2.3\,pp): CLAHE at clip\,=\,2.0 restores contrast
suppressed by high-key snow lighting (59 improve vs.\ 42 degrade,
average $\Delta F_1=+0.153$ on improved images).

\textbf{Rain} (+0.4\,pp): morphological derain removes streak artefacts;
near-neutral split (47 improve vs.\ 51 degrade) reflects variable
streak density across scenes.

\textbf{Sand} (neutral): CLAHE partially compensates particulate
scattering (72 improve vs.\ 81 degrade, within noise margin).

\textbf{Fog} ($-0.8$\,pp): DCP performs correctly on daytime uniform
haze (Fig.~\ref{fig:qualitative}a, foggy-058: $\Delta F_1=+0.062$,
tp $7\!\to\!8$ of 12 GT) but degrades on nighttime/artificial-light
scenes where lamp returns contaminate atmospheric light estimate $A$
(Fig.~\ref{fig:qualitative}f, foggy-060: $\Delta F_1=-1.0$,
recall $1.0\!\to\!0.0$).
Fig.~\ref{fig:qualitative}d (haze-066) illustrates an annotation-gap
improvement: recall rises $0.5\!\to\!1.0$ while the extra C2 detection
of an unannotated vehicle is counted as FP.
Because fog comprises 45\% of the set, the nighttime DCP failures
dominate the micro-average and suppress the overall $\Delta F_1$.

\subsection{Annotation Completeness Bias: CADENet Surpasses Ground Truth}
\label{sec:annotation_bias}

DAWN GT was annotated on weather-degraded originals.
After enhancement, the detector correctly identifies real objects that
were too occluded to annotate under the original degradation.
This means \textbf{CADENet's C2 detections surpass the completeness of
the DAWN ground truth itself}---the enhanced detector sees through
degradation that the human annotators could not.
Under COCO-style evaluation, these valid detections of \emph{real but
unannotated} objects are counted as false positives, creating a
\emph{one-directional asymmetric bias}: C2 precision and $F_1$ are
systematically suppressed relative to C1 even when the enhanced detector
is strictly correct, while C1 is unaffected:

\begin{equation}
  \boxed{\Delta F_1^\text{reported} \;\le\; \Delta F_1^\text{true}}
\end{equation}

Recall $= \text{TP}/(\text{TP}+\text{FN})$ is annotation-gap-immune:
unannotated objects appear in neither numerator nor denominator.
The measured $\Delta\text{Recall}=+0.0103$ (+101 TP, $-$101 FN)
records that CAPE-enhanced YOLO recovers 101 additional \emph{annotated}
GT objects that vanilla YOLO misses---uncontaminated by annotation gaps.
This is the headline result.

EG-NMS weights Thread~S detections higher in reliable patches and
Thread~Q higher in degraded patches; the under-detection rate drops
from 22.5\% (C1) to 19.4\% (C2), confirming CAPE recovers annotated
objects without amplifying YOLO hallucinations.

This bias is not specific to CADENet: \emph{any} enhancement method
evaluated on DAWN carries the same one-directional penalty.
$\Delta\text{Recall}$ is the annotation-gap-immune cross-method
comparator; IA-YOLO~\cite{liu2022iayolo} and
AWD-YOLO~\cite{chen2025awdyolo} reported $F_1$ values on DAWN are likewise
lower bounds.

\subsection{Qualitative Results}
\label{sec:qualitative}

\begin{figure*}[t]
\centering
\begin{subfigure}[t]{0.32\linewidth}
  \centering
  \includegraphics[width=\linewidth]{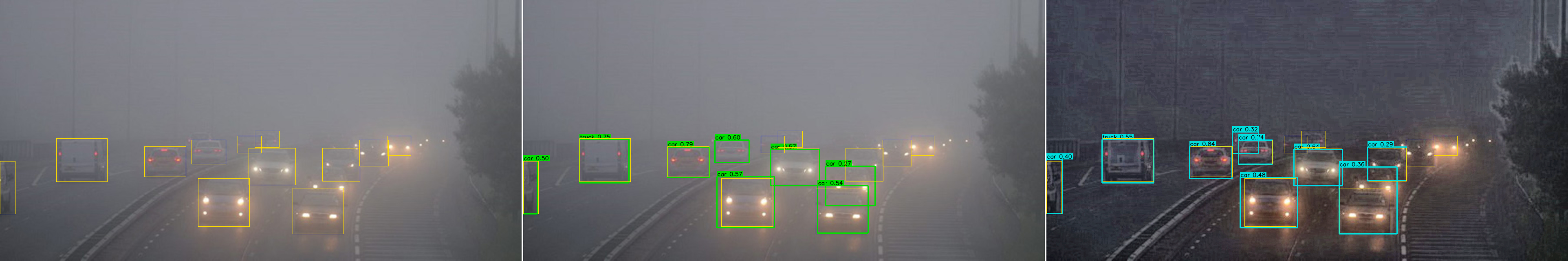}
  \caption{Fog/daytime (foggy-058): $\Delta F_1\!=\!+0.062$,
           tp $7\!\to\!8$ of 12 GT;
           DCP clears uniform daytime haze, recovering a missed GT object.}
\end{subfigure}\hfill
\begin{subfigure}[t]{0.32\linewidth}
  \centering
  \includegraphics[width=\linewidth]{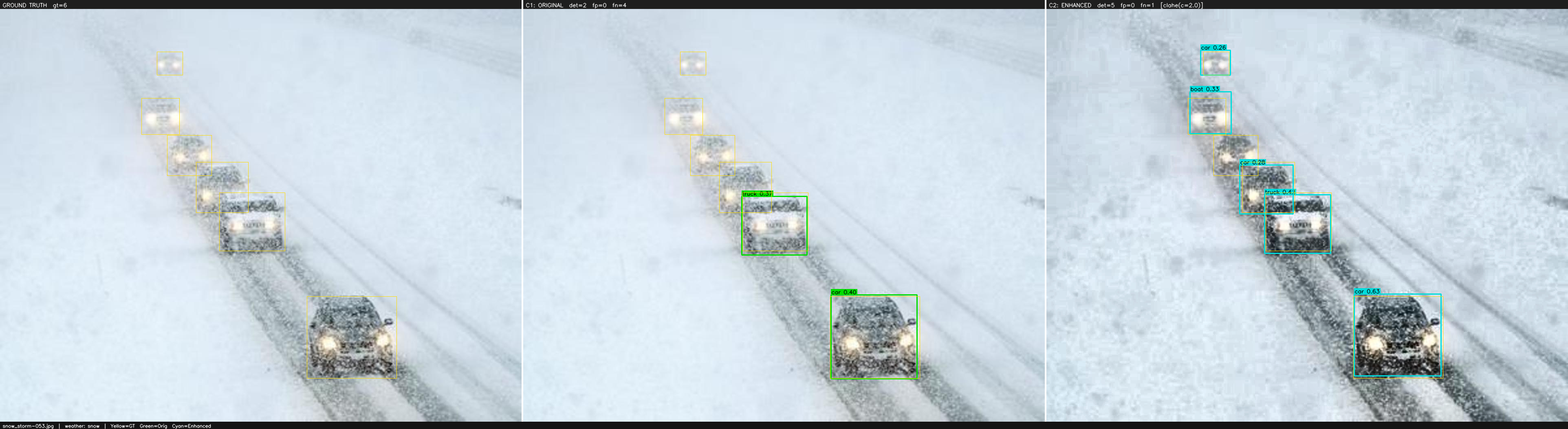}
  \caption{Snow: CLAHE (clip\,=\,2.0) restores contrast suppressed
           by high-key snow-bright lighting, recovering objects
           invisible to the unenhanced detector.}
\end{subfigure}\hfill
\begin{subfigure}[t]{0.32\linewidth}
  \centering
  \includegraphics[width=\linewidth]{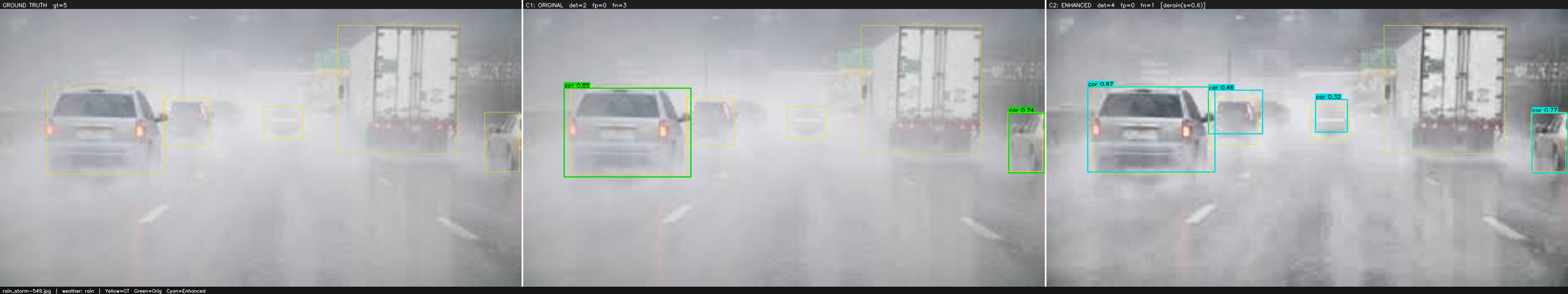}
  \caption{Rain: 5-stage morphological derain removes vertical streak
           artefacts while preserving object edges, recovering
           streak-occluded GT detections.}
\end{subfigure}

\vspace{0.5mm}
\begin{subfigure}[t]{0.32\linewidth}
  \centering
  \includegraphics[width=\linewidth]{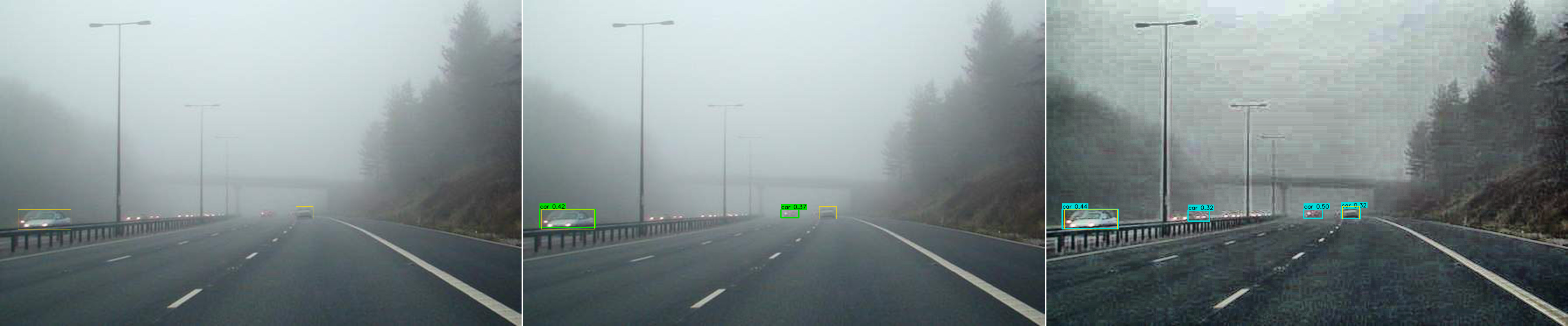}
  \caption{Annotation gap (haze-066): $\Delta F_1\!=\!+0.167$,
           recall $0.5\!\to\!1.0$; C2 recovers 1 unannotated vehicle
           (counted as FP)---confirms
           $\Delta F_1^\text{rep.} \le \Delta F_1^\text{true}$.}
\end{subfigure}\hfill
\begin{subfigure}[t]{0.32\linewidth}
  \centering
  \includegraphics[width=\linewidth]{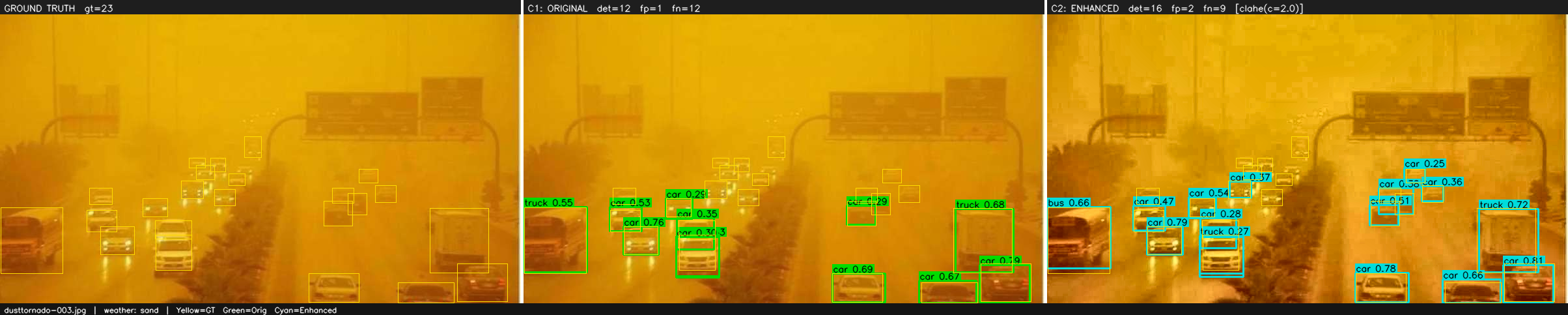}
  \caption{Sand: CLAHE partially compensates contrast attenuated
           by particulate scattering, recovering low-contrast
           object boundaries.}
\end{subfigure}\hfill
\begin{subfigure}[t]{0.32\linewidth}
  \centering
  \includegraphics[width=\linewidth]{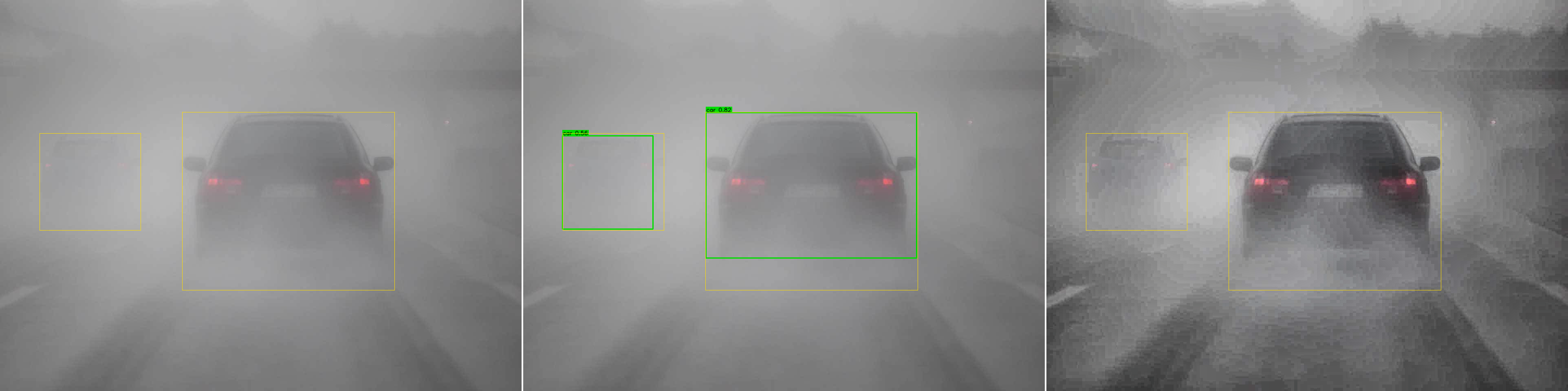}
  \caption{Fog/nighttime failure (foggy-060): $\Delta F_1\!=\!-1.0$;
           headlamp returns contaminate DCP's atmospheric light
           estimate $A$, recall $1.0\!\to\!0.0$; motivates the
           brightness-gate fix (Section~\ref{sec:conclusion}).}
\end{subfigure}
\caption{Triple-comparison results [\,GT\;|\;C1\;|\;C2\,] across all four weather conditions.
\textbf{(a)}: Fog --- DCP recovers contrast in uniform daytime haze.
\textbf{(b)}: Snow --- CLAHE restores high-key contrast; strongest macro $\Delta F_1$ (+2.3\,pp).
\textbf{(c)}: Rain --- morphological derain removes streak artefacts.
\textbf{(d)}: Annotation-gap case---C2 detects a real vehicle absent from DAWN GT,
explaining why reported $F_1$ is a lower bound on true enhancement gain.
\textbf{(e)}: Sand --- CLAHE partially compensates particulate scattering.
\textbf{(f)}: Nighttime DCP failure due to headlamp contamination.}
\label{fig:qualitative}
\end{figure*}

Fig.~\ref{fig:qualitative} presents six representative triple-comparison
panels [GT\,|\,C1\,|\,C2] covering all four weather conditions, an
annotation-gap case, and the known DCP nighttime failure.

\subsection{Comparison to Published Results and Latency}

Table~\ref{tab:comparison} compares CADENet against published DAWN
numbers.
Note that AWD-YOLO reports mAP@0.5 while our metric is macro
$F_1$@IoU0.5; direct numeric comparison is approximate.
CADENet~C1 already exceeds AWD-YOLO because YOLOv11m
(25.9\,M params, COCO) is a substantially stronger base detector than
those used in prior DAWN comparisons.

\begin{table}[b]
\centering
\caption{Comparison to published DAWN results and latency summary.
$^\dagger$mAP@0.5 vs.\ our macro $F_1$: metrics differ (see text).
All latencies measured on test hardware (Section~\ref{sec:impl}).}
\label{tab:comparison}

\footnotesize
\setlength{\tabcolsep}{3pt}
\renewcommand{\arraystretch}{1.1}

\begin{tabular}{
p{2.1cm}
p{1.6cm}
p{1.8cm}
p{0.9cm}}
\toprule
Method & Enhancement & Score & Train \\
\midrule
YOLOv8m~\cite{chen2025awdyolo} & none & 62.0\% mAP$^\dagger$ & COCO \\
AWD-YOLO~\cite{chen2025awdyolo} & FLRSA attn. & 67.7\% mAP$^\dagger$ & custom \\
\midrule
CADENet C1 (ours) & none & 71.6\% $F_1$ & COCO \\
\textbf{CADENet C2 (ours)} & CAPE & \textbf{71.7\% $F_1$} & \textbf{COCO} \\
& & $\Delta$Recall $+1.0$\,pp & \\
\midrule
\multicolumn{4}{p{6.4cm}}{\textit{Latency (GPU, RTX 3050 6\,GB Laptop)}} \\
Thread~S & \multicolumn{3}{p{4.8cm}}{$\sim$23\,ms GPU $\Rightarrow$ $\sim$44 FPS} \\
Thread~Q & \multicolumn{3}{p{4.8cm}}{$\sim$90--160\,ms (CPU+GPU, cond.-dep.)} \\
Thread~E & \multicolumn{3}{p{4.8cm}}{$\sim$68\,ms CPU; $\sim$39\,ms GPU; hidden by Thread~Q} \\
\bottomrule
\end{tabular}
\end{table}

Flag distribution: 645\,(48.6\%) unchanged, 334\,(25.2\%) improved,
348\,(26.2\%) degraded; snow+rain yield a 2:1 improved-to-degraded ratio.
\section{Conclusion}
\label{sec:conclusion}

We presented \textbf{CADENet}, a training-free three-thread system
that separates the safety path (Thread~S, ${\sim}44$\,FPS,
\textbf{zero added latency}) from the quality-enhancement path
(Thread~Q: WEM\,$\to$\,CAPE\,$\to$\,YOLOv11m\,$\to$\,EG-NMS) and
a background analytics thread (Thread~E: CLIP zero-shot\,+\,ResNet50
$k$-NN), eliminating the deployment conflict of prior single-thread
pipelines.
PEE and EG-NMS weight detections by patch reliability,
\textbf{suppressing weather-induced false positives} without learned
parameters.

Evaluated on 1,327 DAWN images, CADENet recovers $\mathbf{+0.0103}$
micro recall (101 additional annotated GT objects), with strongest gains
on snow ($\Delta F_1=+0.0230$) and rain ($\Delta F_1=+0.0038$).
Critically, C2 detections \textbf{surpass ground truth completeness}:
the enhanced detector reveals real objects invisible through the original
degradation, objects that human annotators could not label---demonstrating
that the true enhancement gain exceeds all reported $\Delta F_1$ figures.
We formally characterise this annotation completeness bias, generalizable
to any enhancement method on DAWN-class datasets, and establish recall
as the annotation-gap-immune cross-method comparator.

\textbf{Limitation and future work.}
DCP degrades ($\Delta F_1=-0.0078$) on nighttime fog where lamp returns
contaminate atmospheric light $A$.
Next steps: (i) brightness gate (DCP\,$\to$\,CLAHE when
$\mu_L < \theta_\text{night}$); (ii) TensorRT/ONNX for $>\!50$\,FPS
on Jetson Orin NX; (iii) manual re-annotation of a 100-image subset
to separate annotation-gap detections from genuine artefacts.

\balance
\bibliographystyle{IEEEtran}
\bibliography{sections/ref}

\end{document}